\documentclass[runningheads]{llncs}
\usepackage[rgb,x11names]{xcolor}
\usepackage{amssymb}
\usepackage{amsmath}
\usepackage{subfig}
\usepackage{hyperref}
\hypersetup{colorlinks=true,citecolor=blue}
\usepackage{graphicx}
\usepackage[]{algorithm2e}
\usepackage{mathrsfs}
\usepackage{bbding}
\usepackage{rotating}
\usepackage{multirow}
\usepackage{marginnote}
\usepackage{overpic}
\usepackage{colortbl}
\usepackage[labelfont=bf]{caption}
\usepackage{bbding}
\usepackage{booktabs}

\newcommand{\secref}[1]{$\S$ \ref{#1}}

\newcommand{\figref}[1]{Fig. \ref{#1}}
\newcommand{\tabref}[1]{Tab. \ref{#1}}

\newcommand{\supp}[1]{\textcolor{magenta}{#1}}

\definecolor{mygray}{gray}{.92}
\def\etal{{\em et al.}}
\def\ie{\emph{i.e.}}
\def\eg{\emph{e.g.}}
\def\etc{\emph{etc}}
\def\etal{{\em et al.~}}

\newcommand{\tabincell}[2]{\begin{tabular}{@{}#1@{}}#2\end{tabular}}
\def\ourmodel{\textit{PNS-Net}}
\newcommand{\myPara}[1]{\noindent\textbf{#1~}}

\makeatletter
\newcommand\figcaption{\def\@captype{figure}\caption}
\newcommand\tabcaption{\def\@captype{table}\caption}
\makeatother

\newcommand\blfootnote[1]
{%
	\begingroup
	\renewcommand\thefootnote{}
	\footnotetext{#1}%
	\endgroup
}

\graphicspath{{./Imgs/}}
\DeclareGraphicsExtensions{.jpg,.pdf,.png}

\begin{document}
\title{Progressively Normalized Self-Attention Network for Video Polyp Segmentation}
\author{
Ge-Peng~Ji\inst{1,2}\and 
Yu-Cheng~Chou\inst{2}\and 
Deng-Ping~Fan\inst{1,}\Envelope\\ 
Geng~Chen\inst{1}\and 
Huazhu~Fu\inst{1}\and 
Debesh~Jha\inst{3}\and 
Ling~Shao\inst{1} 
} 

%
%

\institute{
$^1$ Inception Institute of AI (IIAI)
$^2$ Wuhan University
$^3$ SimulaMet\\
{\tt \small dengpfan@gmail.com}
}

%
\maketitle              
\begin{abstract}
Existing video polyp segmentation\blfootnote{G.-P. Ji and Y.-C. Chou contributed equally. Code: \href{http://dpfan.net/pranet/}{http://dpfan.net/pranet/}}(VPS) models typically employ convolutional neural networks (CNNs) to extract features. 
However, due to their limited receptive fields, CNNs cannot fully exploit the global temporal and spatial information in successive video frames, resulting in false positive segmentation results.
In this paper, we propose the novel \textbf{\ourmodel} (Progressively Normalized Self-attention Network), which can efficiently learn representations from polyp videos with real-time speed (\textbf{$\sim$140fps}) on a single RTX 2080 GPU and no post-processing.  
Our \ourmodel~is based solely on a basic normalized self-attention block, equipping with recurrence and CNNs entirely.  
Experiments on challenging VPS datasets demonstrate that the proposed \ourmodel~achieves state-of-the-art performance.
%
We also conduct extensive experiments to study the effectiveness of the channel split, soft-attention, and progressive learning strategy.
We find that our \ourmodel~works well under different settings, making it a promising solution to the VPS task.
%
%

\end{abstract}
%
%
%


\section{Introduction}
Early diagnosis of colorectal cancer (CRC) plays a vital role in improving the survival rate of CRC patients. 
In fact, the survival rate in the first stage of CRC is over 95\%, decreasing to below 35\% in the fourth and fifth stages \cite{bernal2012towards}.
Currently, colonoscopy is widely adopted in clinical practice and has become a standard method for screening CRC.
During the colonoscopy, physicians visually inspect the bowel with an endoscope to identify polyps, which can develop into CRC if left untreated.
In practice, colonoscopy is highly dependent on the physicians' level of experience and suffers from a high polyp miss rate \cite{puyal2020endoscopic}.
These limitations can be resolved with automatic polyp segmentation techniques, which segment polyps from colonoscopy images/videos without intervention from physicians.
However, accurate and real-time polyp segmentation is a challenging task due to the low boundary contrast between a polyp and its surroundings and the large shape variation of polyps~\cite{fan2020pra}.

Significant efforts have been dedicated to overcoming these challenges. 
In early studies, learning-based methods turned to handcrafted features~\cite{mamonov2014automated,tajbakhsh2015automated}, such as color, shape, texture, appearance, or some combination. 
These methods train a classifier to separate the polyps from the background. 
However, they usually suffer from low accuracy due to the limited representation capability of handcrafted features in depicting heterogeneous polyps, as well as the close resemblance between polyps and hard mimics~\cite{yu2016integrating}.
In more recent studies, deep learning methods have been used for polyp segmentation~\cite{yu2016integrating,zhang2018polyp}. 
Although these methods have made some progress, they only use bounding boxes to detect polyps, and therefore cannot accurately locate the boundaries. 
To solve this, Brandao~\etal~\cite{brandao2017fully} adopted a fully convolutional networks (FCN) with a pre-trained model to recognize and segment polyps. 
Later, Akbari~\etal~\cite{akbari2018polyp} introduced a modified FCN to increase the accuracy of polyp segmentation.
Inspired by the success of UNet~\cite{ronneberger2015u} in biomedical image segmentation, UNet++~\cite{zhou2018unetplus} and ResUNet~\cite{jha2019resunetplus} are employed for polyp segmentation and achieved good results.
Some methods also focus on area-boundary constraints.
For instance, Psi-Net~\cite{murugesan2019psi} makes use of polyp boundary and area information simultaneously. 
Fang~\etal~\cite{fang2019selective} introduced a three-step selective feature aggregation network. 
ACSNet~\cite{zhang2020adaptive} utilized an adaptive context selection based encoder-decoder framework. 
Zhong~\etal~\cite{zhong2020polypseg} propose a context-aware network based on adaptive scale and global semantic context. 
Introduced more recently, the current golden standard for image polyp segmentation, PraNet~\cite{fan2020pra}, applies area and boundary cues in a reverse attention module, achieving the cutting-edge performance. 
However, these methods have only been trained and evaluated on still images and focus on static information, ignoring the temporal information in endoscopic videos which can be exploited for better results. 
To this end, Puyal~\etal~\cite{puyal2020endoscopic} propose a hybrid 2D/3D CNN architecture. 
Their model aggregates spatial and temporal correlations and achieves better segmentation results. 
However, the spatial correlation between frames is restricted by the size of the kernel, preventing the accurate segmentation of fast videos. 

Recently, the self-attention network~\cite{wang2018non} has shown superior performance in computer vision tasks such as video object segmentation~\cite{gu2020pyramid}, image super-resolution~\cite{yang2020learning}, and others.
Inspired by this, in this paper, we propose a novel self-attention framework, called the \textbf{P}rogressively \textbf{N}ormalized \textbf{S}elf-attention \textbf{Net}work (\textbf{\ourmodel}), for the video polyp segmentation (VPS) task. 
Our contributions are as follows:
\begin{itemize}
    \item[$\bullet$] Different from existing CNN-based models, the proposed \ourmodel~framework is a self-attention model for VPS, introducing a new perspective for addressing this task. 
    
    \item[$\bullet$] To fully utilize the temporal and spatial cues, we propose a simple normalized self-attention (NS) block. The NS block is flexible (backbone-free) and efficient, enabling it to easily be embedded into current CNN-based encoder-decoder architectures for better performance.
    \item[$\bullet$] We evaluate the proposed \ourmodel~on challenging VPS datasets and compare it with two classical methods (\ie, UNet~\cite{ronneberger2015u} and UNet++~\cite{zhou2018unetplus}) and three cutting-edge models (\ie, ResUNet~\cite{jha2019resunetplus}, ACSNet~\cite{zhang2020adaptive}, and PraNet~\cite{fan2020pra}). Experimental results show that \ourmodel~achieves state-of-the-art performance with real-time speed. All the training data, models, results, and evaluation tools will be released to advance the development of this field.
\end{itemize}


\section{Method}

\begin{figure}[!t]
  \centering
  \includegraphics[width=0.8\linewidth]{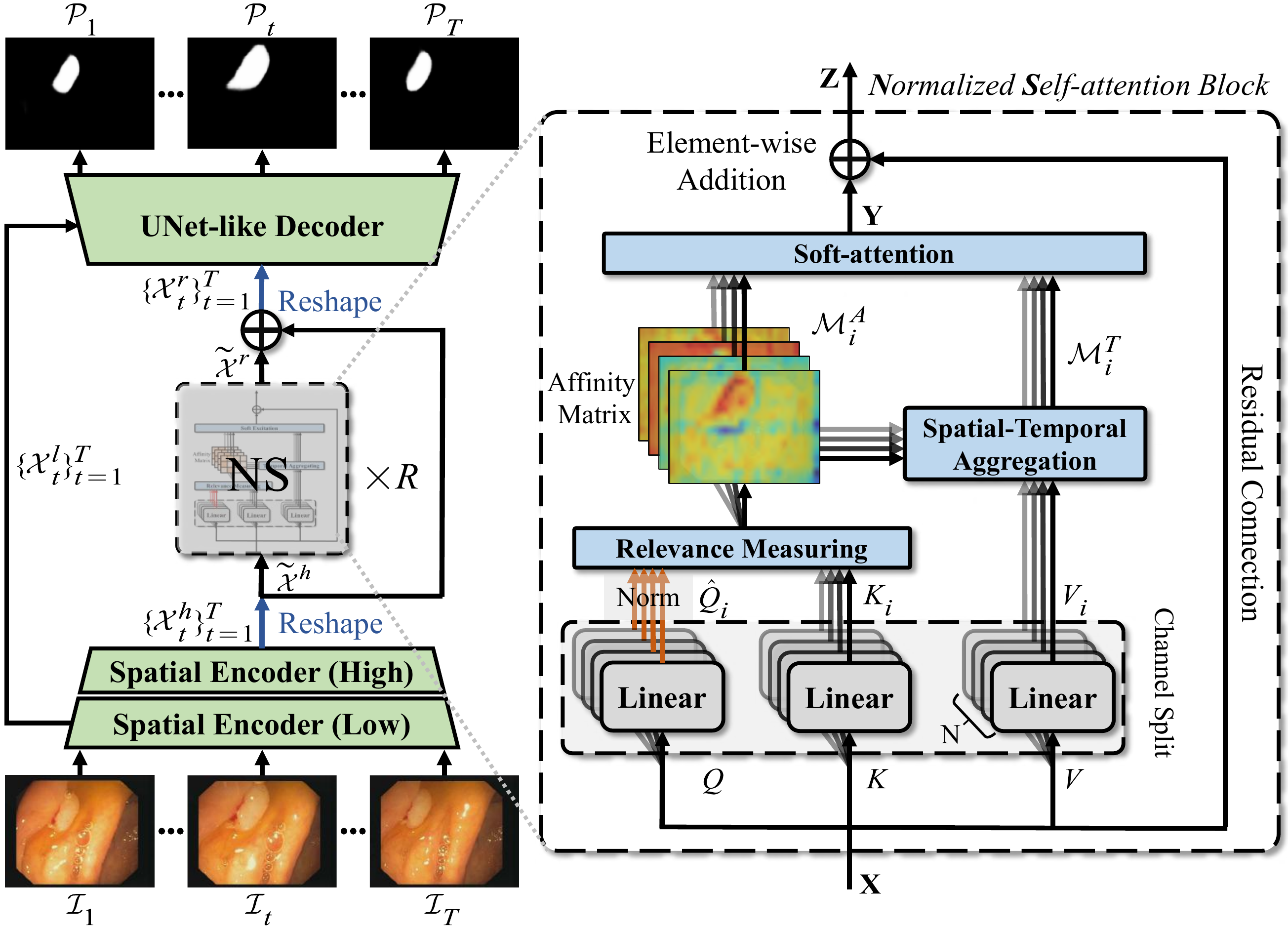}
  \caption{
    Pipeline of the proposed \ourmodel, including the normalized self-attention block (see~\secref{sec:NS}) with a stacked ($\times$R) learning strategy (see~\secref{sec:pipeline}).
  }\label{fig:main_framework}
\end{figure}

\subsection{Normalized Self-attention (NS)}\label{sec:NS}
\myPara{Motivation.} 
Recently, the self-attention mechanism~\cite{wang2018non} has been widely exploited in many popular computer vision tasks. 
However, in our initial studies, we found that introducing the original self-attention mechanism to the VPS task does not achieve satisfactory results (\ie, high accuracy and speed).

\myPara{Analysis.} For the VPS task, multi-scale polyps move at various speeds. Thus, dynamically updating the receptive field of the network is important. 
Further, the self-attention, such as the non-local network \cite{wang2018non}, incurs a high computational and memory cost, which limits the inference speed for our fast and dense prediction task.
Motivated by the recent video salient object detection model~\cite{gu2020pyramid}, we utilize the \textbf{channel split}, \textbf{query-dependent}, and \textbf{normalization} rules to reduce the computational cost and improve the accuracy, respectively. 

\textit{Channel Split Rule.} Specifically, given an input feature (\ie, $\mathbf{X} \in \mathbb{R}^{T \times H \times W \times C}$) extracted from $T$ video frames with a size of $H \times W$ and $C$ channels, we first utilize three linear embedding functions $\theta(\cdot)$, $\phi(\cdot)$, and $g(\cdot)$ to generate the corresponding attention features, which are implemented by a $1\times1\times1$ convolutional layer~\cite{wang2018non}. 
This can be expressed as:
\begin{equation}
    Q = \theta(\mathbf{X}), K = \phi(\mathbf{X}), V = g(\mathbf{X}).
\end{equation}
Then we split each attention feature $\{ Q,K,V \} \in \mathbb{R}^{T \times H \times W \times C}$ into $N$ groups along the channel dimension and generate query, key, and value features, \ie, $\{ Q_i,K_i,V_i \} \in \mathbb{R}^{T \times H \times W \times \frac{C}{N}}$, where $i = \{ 1, 2, \cdots, N \}$.

\textit{Query-Dependent Rule.}
To extract the spatial-temporal relationship between successive video frames, we need to measure the similarity between query features $Q_i$ and key features $K_i$.
Inspired by~\cite{gu2020pyramid}, we introduce $N$ relevance measuring (\ie, query-dependent rule) blocks 
to compute the spatial-temporal affinity matrix for the \textit{constrained neighborhood} of the target pixel.
Rather than computing the response between a query position and the feature at all positions, as done in~\cite{wang2018non}, the relevance measuring block can capture more relevance 
regarding the target object within $T$ frames. 
More specifically, given a sliding window with fixed kernel size $k$ and dilation rate $d_i=2i-1$, 
we get the corresponding constrained neighborhood in $K_i$ for query pixel $\mathbf{X}^q$ of $Q_i$ in position $(x,y,z)$, which can be obtained by a sampling function $\mathcal{F}^S$.
This is computed by:
\begin{equation}
   \mathcal{F}^S \langle \mathbf{X}^q , K_i \rangle \in \mathbb{R}^{T(2k+1)^2 \times \frac{C}{N}} = \Sigma_{m=x-kd_i}^{x+kd_i} \Sigma_{n=y-kd_i}^{y+kd_i} \Sigma_{t=1}^{T} K_i(m, n, t),
\end{equation}
where $1 \leq x \leq H$, $1 \leq y \leq W$, and $1 \leq z \leq T$.
Thus, the size of the constrained neighborhood depends on the various spatial-temporal receptive fields with different kernel size $k$, dilation rate $d_i$, and frame number $T$, respectively.

\textit{Normalization Rule.}
However, the internal covariate shift problem~\cite{guo2020normalized} exists in the feed-forward of input $Q_i$, incurring that the layer parameters cannot dynamically adapt the next mini-batch.
Therefore, we maintain a fixed distribution for $Q_i$ via: 
\begin{equation}
    \hat{Q}_i = \texttt{Norm} (Q_i),
\end{equation}
where \texttt{Norm} is implemented by layer normalization~\cite{ba2016layer} \textit{along temporal dimension.}

\myPara{Relevance Measuring.} Finally, the affinity matrix is computed as:
\begin{equation}\label{equ:relevance_measuring}
    \mathcal{M}^A_i \in \mathbb{R}^{THW\times T(2k+1)^2} = \texttt{Softmax}(\frac{\hat{Q}_i \mathcal{F}^S \langle \hat{\mathbf{X}}^q , K_i \rangle^{\mathbf{T}} }{\sqrt{C/N}}),~\text{when}~\hat{\mathbf{X}}^q \in \hat{Q}_i,
\end{equation}
where $\sqrt{C/N}$ is a scaling factor to balance the multi-head attention~\cite{vaswani2017attention}.

\myPara{Spatial-Temporal Aggregation.}
Similar to relevance measuring, we also compute the spatial-temporally aggregated features $\mathcal{M}^T_i$ within the constrained neighborhood during temporal aggregation.
This can be formulated as:
\begin{equation}\label{equ:s_t_aggregating}
    \mathcal{M}^T_i \in \mathbb{R}^{THW \times \frac{C}{N}} = \mathcal{M}^A_i \mathcal{F}^S \langle \mathbf{X}^a, V_i \rangle,~\text{when}~\mathbf{X}^a \in \mathcal{M}^A_i,
\end{equation}

\myPara{Soft-Attention.}
We use a soft-attention block to synthesize features from the group of 
affinity matrices $\mathcal{M}^A_i$ and aggregated features $\mathcal{M}^T_i$.
During the synthesis process, relevant spatial-temporal patterns should be enhanced 
while less relevant ones should be suppressed.
We first concatenate a group of affinity matrices $\mathcal{M}^A_i$ along the channel dimension to generate $\mathcal{M}^A$.
Thus, the soft-attention map $\mathcal{M}^S$ is computed by:
\begin{equation}
    \mathcal{M}^S \in \mathbb{R}^{THW \times 1}  \leftarrow \texttt{max} \mathcal{M}^A,~\text{when}~\mathcal{M}^A \in \mathbb{R}^{THW \times  T(2k+1)^2N},
\end{equation}
where the $\texttt{max}$ function computes the channel-wise maximum value.
Then we concatenate a group of the spatial-temporally aggregated features $\mathcal{M}^T_i$ along the channel dimension to generate $\mathcal{M}^T$.

\myPara{Normalized Self-attention.}
Finally, our NS block can be computed as:
\begin{equation}
    \mathbf{Z} \in \mathbb{R}^{T \times {H} \times {W} \times C} = \mathbf{X} + \mathbf{Y} = \mathbf{X} + (\mathcal{M}^T \mathbf{W}_T) \circledast \mathcal{M}^S,
\end{equation}
where $\mathbf{W}_T$ is the learnable weight and $\circledast$ is the channel-wise Hadamard product.

\subsection{Progressive Learning Strategy}\label{sec:pipeline}

\myPara{Encoder.}
For fair comparison, we use the same backbone (\ie, Res2Net-50) as in PraNet~\cite{fan2020pra}.
%
Given a polyp video clip with $T$ frames as input (\ie, $\{ \mathcal{I} \}_{t=1}^{T} \in \mathbb{R}^{H' \times W' \times 3}$),
we first feed it into a spatial encoder to extract two spatial features from the conv3\_4 and conv4\_6 layers, respectively.
To alleviate the computational burden, we adopt an RFB-like~\cite{liu2018receptive} module to reduce the feature channel.
Thus, we generate two spatial features, including low-level (\ie, $\{ \mathcal{X}^{l}_t\}_{t=1}^T \in \mathbb{R}^{H^{l} \times W^{l} \times C^{l}}$) and high-level (\ie, $\{ \mathcal{X}^{h}_t \}_{t=1}^T \in \mathbb{R}^{H^{h} \times W^{h} \times C^{h}}$)\footnote{We set $H^{l}=\frac{H'}{4}$, $W^{l}=\frac{W'}{4}$, $C^{l}=24$, $H^{h}=\frac{H'}{8}$, $W^{h}=\frac{W'}{8}$, and $C^{h}=32$.}.

\myPara{Progressively Normalized Self-attention (PNS).}
Most attention strategies aim to refine candidate features, such as first-order~\cite{fan2020pra} and second-order~\cite{wang2018non,vaswani2017attention} functions.
%
%
As such, the strong semantic information in high-level features might be diffused gradually during the forward pass of the network.
To alleviate this, we introduce a progressive residual learning strategy in our NS block.
Specifically, we first reshape the corresponding high-level features $\{ \mathcal{X}^{h}_t \}_{t=1}^T$ of consecutive input frames into a temporal feature, which can be viewed as a four-dimensional tensor (\ie, $\widetilde{\mathcal{X}}^{h} \in \mathbb{R}^{T \times H^{h} \times W^{h} \times C^{h}}$).
Then we refine $\widetilde{\mathcal{X}}^{h}$ via stacked normalized self-attention in a progressive manner:
\begin{equation}
    \widetilde{\mathcal{X}}^{r} \in \mathbb{R}^{T \times H^{h} \times W^{h} \times C^{h}} = \texttt{NS}^{\times R} ( \widetilde{\mathcal{X}}^{h}) = \texttt{NS}^{\times R} (\mathcal{F}^R (\{ \mathcal{X}^{h}_t \}_{t=1}^T)),
\end{equation}
where $\texttt{NS}^{\times R}$ means that $R$ normalized self-attention blocks are stacked in the refinement process.
$\mathcal{F}^R$ is the reshaping function for the temporal dimension.
To allow this block to easily be plugged into pre-trained networks, the commonly adopted solution is to add a residual learning process.
Finally, the refined spatial-temporal feature is generated by:
\begin{equation}
    \{ \mathcal{X}^{r}_t \}_{t=1}^T \in \mathbb{R}^{{H}^{h} \times {W}^{h} \times C^{h}}  = \mathcal{F}^R (\widetilde{\mathcal{X}}^h + \widetilde{\mathcal{X}}^r).
\end{equation}
%

\myPara{Decoder and Learning Strategy.}
We combine the low-level feature $\{ \mathcal{X}^{l}_t \}_{t=1}^T$ from the spatial decoder and the spatial-temporal feature $\{ \mathcal{X}^{r}_t \}_{t=1}^T$ from the PNS block via a two-stage UNet-like decoder $\mathcal{F}^D$. 
Thus, the output of our method is computed by $\{ \mathcal{P}_t \}_{t=1}^T = \mathcal{F}^D(\{ \mathcal{X}^{l}_t \}_{t=1}^T, \{ \mathcal{X}^{r}_t \}_{t=1}^T )$.
We adopt the standard \textit{cross-entropy} loss function in the learning process. 

\section{Experiments}


\subsection{Implementation Details}

\myPara{Datasets.} 
We adopt four widely used polyp datasets in our experiments, including image-based (\ie, Kvasir~\cite{jha2020kvasir}) and video-based (\ie,  CVC-300~\cite{bernal2012towards}, CVC-612~\cite{bernal2015wm}, and ASU-Mayo~\cite{tajbakhsh2015automated}) ones.
Kvasir is a large-scale and challenging dataset, which consists of 1,000 polyp images with fully annotated pixel-level ground truths (GTs).
The whole Kvasir is used for training.
%
%
ASU-Mayo contains 10 negative video samples from normal subjects and 10 positive samples from patients.
We only adopt the positive part for training.
Following the same protocol as~\cite{bernal2012towards,bernal2015wm}, we split the videos from CVC-300 (12 clips) and CVC-612 (29 clips) into $60\%$ for training, $20\%$ for validation, and $20\%$ for testing.

\myPara{Training.} 
Due to the limited video training data, we try to fully utilize large-scale image data to capture more appearances of the polyp and scene. Thus, we train our model in two steps:
\textit{i) Pre-training phase.} We remove the normalized self-attention (NS) block from \ourmodel~and pre-train the static backbone using an image-based polyp dataset (\ie, Kvasir~\cite{jha2020kvasir}) and the training set of video-based polyp datasets (\ie,  CVC-300~\cite{bernal2012towards}, CVC-612~\cite{bernal2015wm}, and ASU-Mayo~\cite{tajbakhsh2015automated}).
The initial learning rate of the Adam algorithm and the weight decay are both 1e{-4}.
The static part of our \ourmodel~convergences after 100 epochs.
\textit{ii) Fine-tuning phase.} We plug the NS block into 
our \ourmodel~and fine-tune the whole network 
using the video polyp datasets, including the ASU-Mayo and the training sets of CVC-300 and CVC-612.
%
%
We set the number of attention groups $N=4$ and the number of stacked normalized self-attention blocks $R=2$, along with a kernel size of $k=3$.
The initial learning is set to 1e{-4}, and the whole model is fine-tuned over one epoch.
In this way, although the densely labeled VPS data is scarce, our \ourmodel~still achieves good generalization performance.

\myPara{Testing and Runtime.}
%
To test the performance of our \ourmodel,
we validate it on challenging datasets, including the test set of CVC-612 (\ie, CVC-612-T), the validation set of CVC-612 (\ie,  CVC-612-V), and the test/validation set of CVC-300 (\ie, CVC-300-TV).
During inference, we sample $T$=5 frames from a polyp clip and resize them to 256$\times$448 as the input.
For final prediction, we use the output $\mathcal{P}_t$ of the network followed by a \textit{sigmoid} function.
Our \ourmodel~achieves a speed of $\sim$140fps on a single RTX 2080 GPU without any post-processing (\eg, CRF~\cite{krahenbuhl2011efficient}).
The speeds of the compared methods are listed in \tabref{tab:ModelScore}.

\subsection{Evaluation on Video Polyp Segmentation}

\myPara{Baselines.} 
We re-train five cutting-edge polyp segmentation baselines (\ie, UNet~\cite{ronneberger2015u}, UNet++~\cite{zhou2018unetplus}, ResUNet~\cite{jha2019resunetplus}, ACSNet~\cite{zhang2020adaptive}, and PraNet~\cite{fan2020pra}) with the same data used by our \ourmodel, under their default settings, for fair comparison.


\myPara{Metrics.} 
The metrics used included:
(1) maximum Dice (maxDice), which measures the similarity between two sets of data;
(2) maximum specificity (maxSpe), which refers to the percentage of the samples that are negative and are judged as such;
(3) maximum IoU (maxIoU), which measures the overlap between two masks;
(4) S-measure~\cite{fan2017structure} ($S_{\alpha}$), which evaluates region- and object-aware structural similarity;
(5) enhanced-alignment measure~\cite{21SC-Emeasure} ($E_\phi$), which measures pixel-level matching and image-level statistics;
and (6) mean absolute error ($M$), which measures the pixel-level error between the prediction and GT.

\begin{figure}[t!]
  \centering
  \includegraphics[width=\columnwidth]{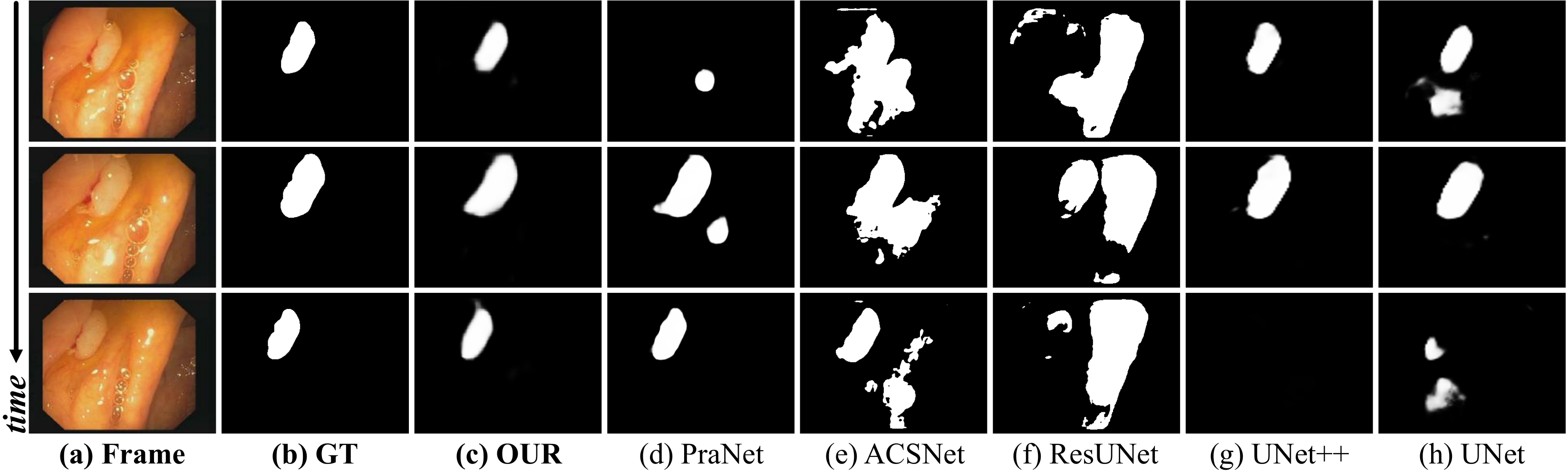}
  \caption{
  Qualitative results on CVC-612-T~\cite{bernal2015wm}. For more visualization results please refer to the \supp{supplementary material (\ie, PDF file and videos)}. 
  }\label{fig:front_figure}
\end{figure}

\myPara{Qualitative Comparison.}
In~\figref{fig:front_figure}, We provide the polyp segmentation results of our \ourmodel~on CVC-612-T. Our model can accurately locate and segment polyps in many difficult situations, such as different sizes, homogeneous areas, different textures, \etc.

\begin{table}[t!]
  \centering
  \scriptsize
  \renewcommand{\arraystretch}{0.95}
  \setlength\tabcolsep{4.9pt}
  \caption{Quantitative results on different datasets. 
    }\label{tab:ModelScore}
  \begin{tabular}{rl||ccc|cc|c} 
  \hline \toprule
  & & \multicolumn{3}{c|}{\tabincell{c}{2018$\sim$2019}} &\multicolumn{2}{c|}{\tabincell{c}{2020}} & 2021 \\
  \cline{3-8}
  && UNet &UNet++ &ResUNet & ACSNet & PraNet & \textbf{\ourmodel} \\
  & Metrics & MICCAI~\cite{ronneberger2015u} & TMI~\cite{zhou2018unetplus} & ISM~\cite{jha2019resunetplus} & MICCAI~\cite{zhang2020adaptive} & MICCAI~\cite{fan2020pra} & \textbf{(OUR)} \\
\hline
 & Speed & 108fps   & 45fps   & 20fps   & 35fps   & 97fps   & \textbf{140fps}   \\
\hline
\multirow{6}{*}{\begin{sideways}CVC-300-TV\end{sideways}} 
& maxDice$\uparrow$ & 0.639   & 0.649   & 0.535   & 0.738   & 0.739   & \textbf{0.840}   \\
& maxSpe$\uparrow$ & 0.963   & 0.944   & 0.852   & 0.987   & 0.993   & \textbf{0.996}   \\
& maxIoU$\uparrow$ & 0.525   & 0.539   & 0.412   & 0.632   & 0.645   & \textbf{0.745}   \\
& $S_\alpha\uparrow$ & 0.793   & 0.796   & 0.703   & 0.837   & 0.833   & \textbf{0.909}   \\
& $E_\phi\uparrow$ & 0.826   & 0.831   & 0.718   & 0.871   & 0.852   & \textbf{0.921}   \\
& $M\downarrow$ & 0.027   & 0.024   & 0.052   & 0.016   & 0.016   & \textbf{0.013}   \\
\hline
\multirow{6}{*}{\begin{sideways}CVC-612-V\end{sideways}} 
& maxDice$\uparrow$& 0.725   & 0.684   & 0.752   & 0.804   & 0.869   & \textbf{0.873}\\
& maxSpe$\uparrow$ & 0.971   & 0.952   & 0.939   & 0.929   & 0.983   & \textbf{0.991}\\
& maxIoU$\uparrow$& 0.610   & 0.570   & 0.648   & 0.712   & 0.799   & \textbf{0.800}\\
& $S_\alpha\uparrow$ & 0.826   & 0.805   & 0.829   & 0.847   & 0.915   & \textbf{0.923}\\
& $E_\phi\uparrow$ & 0.855   & 0.830   & 0.877   & 0.887   & 0.936   & \textbf{0.944}\\
& $M\downarrow$ & 0.023   & 0.025   & 0.023   & 0.054   & 0.013   & \textbf{0.012}\\
\hline
\multirow{6}{*}{\begin{sideways}CVC-612-T\end{sideways}} 
& maxDice$\uparrow$ & 0.729   & 0.740   & 0.617   & 0.782   & 0.852   & \textbf{0.860}\\
& maxSpe$\uparrow$ & 0.971   & 0.975   & 0.950   & 0.975   & 0.986   & \textbf{0.992}\\
& maxIoU$\uparrow$ & 0.635   & 0.635   & 0.514   & 0.700   & 0.786   & \textbf{0.795}\\
& $S_\alpha\uparrow$ & 0.810   & 0.800   & 0.727   & 0.838   & 0.886   & \textbf{0.903}\\
& $E_\phi\uparrow$ & 0.836   & 0.817   & 0.758   & 0.864   & \textbf{0.904}   & 0.903\\
& $M\downarrow$ & 0.058   & 0.059   & 0.084   & 0.053   & 0.038   & \textbf{0.038}\\
    \bottomrule
  \end{tabular}
\end{table}

\myPara{Quantitative Comparison.}
Quantitative comparison results are summarized in~\tabref{tab:ModelScore}.
We conduct three experiments on
 test datasets to verify the model's performance.
CVC-300-TV consists of both validation set and test set, which include six videos in total.
CVC-612-V and CVC-612-T each contain five videos.
On CVC-300, where all the baseline methods perform poorly, our \ourmodel~achieves remarkable performance in all metrics and outperforms all SOTA methods by a large margin (max Dice: $\sim$10\%).
On CVC-612-V and CVC-612-T, our \ourmodel~consistently outperforms other SOTAs.

\subsection{Ablation Study}\label{sec:ablation}

\begin{table}[t!]
  \centering
  \scriptsize
  \renewcommand{\arraystretch}{0.6}
  \setlength\tabcolsep{1.8pt}
  \caption{Ablation studies.
    See \secref{sec:ablation} for more details.
    }\label{tab:AblationScore}
  \begin{tabular}{l || ccccc || cccc | cccc} 
  \hline \toprule
    & \multicolumn{5}{c||}{\tabincell{c}{Variants}}  
    & \multicolumn{4}{c|}{\tabincell{c}{CVC-300-TV}} 
    & \multicolumn{4}{c}{\tabincell{c}{CVC-612-T}}
    \\ 
    \cline{2-14}
    No. & Base & N & Soft &Norm & R
    &maxDice$\uparrow$ &maxIoU$\uparrow$ &$S_\alpha\uparrow$ &$E_\phi\uparrow$ 
    &maxDice$\uparrow$ &maxIoU$\uparrow$ &$S_\alpha\uparrow$ &$E_\phi\uparrow$ \\ 
    \hline
\#1 & \checkmark & & & &
& 0.778   & 0.665   & 0.850   & 0.858   
& 0.850   & 0.778   & 0.896   & 0.885     \\	
\hline
\#2 & \checkmark & 1 & & & 1
& 0.755   & 0.650   & 0.865   & 0.844   
& 0.850   & 0.779   & 0.896   & 0.891   \\	
\#3 & \checkmark & 2 & & & 1
& 0.790   & 0.679   & 0.876   & 0.872   
& 0.825   & 0.746   & 0.870   & 0.856   \\	
\#4 & \checkmark & 4 & & & 1
& 0.809   & 0.709   & 0.893   & 0.884   
& 0.834   & 0.760   & 0.881   & 0.867   \\	
\#5 & \checkmark & 8 & & & 1
& 0.763   & 0.663   & 0.867   & 0.842   
& 0.787   & 0.702   & 0.841   & 0.829   \\	
\hline
\#6 & \checkmark & 4 & \checkmark & & 1 
& 0.829   & 0.729   & 0.896   & 0.903   
& 0.852   & 0.784   & 0.895   & 0.897   \\	
\hline
\#7 & \checkmark & 4 & \checkmark & \checkmark & 1 
& 0.827   & 0.732   & 0.897   & 0.898   
& 0.856   & 0.792   & 0.898   & 0.896   \\	
\#8 & \checkmark & 4 & \checkmark & \checkmark & 2
& \textbf{0.840}   & \textbf{0.745}   & \textbf{0.909}   & \textbf{0.921}
& \textbf{0.860}   & \textbf{0.795}   & \textbf{0.903}   & \textbf{0.903} \\
\#9 & \checkmark & 4 & \checkmark & \checkmark & 3 
& 0.737   & 0.609   & 0.793   & 0.751   
& 0.732   & 0.613   & 0.776   & 0.728   \\	
    \bottomrule
  \end{tabular}
\end{table}

\myPara{Effectiveness of Channel Split.}
We investigate the contribution of channel split rule under different scales.
The results are listed in rows \#2 to \#5 in~\tabref{tab:AblationScore}.
We observe that \#4 (N=4) outperforms other settings (\ie, \#2, \#3, and \#5) 
on CVC-300-TV, in all metrics.
This improvement shows that an improper receptive field (RF) harms the ability to excavate temporal information, since a large RF will pay more attention to the global environment rather than local motion information. 
%
On the other hand, when the split number is too small, the model fails to capture multi-scale polyps moving at various speeds.

\myPara{Effectiveness of Soft-attention.}
We further investigate the contribution of the soft-attention mechanism.
As shown in~\tabref{tab:AblationScore}, \#6 is generally better than \#4 with the soft-attention block on CVC-612-T.
%
%
This improvement suggests that introducing the soft-attention block to synthesize the aggregation feature and affinity matrix is necessary for increasing performance.

\myPara{Effectiveness of the Number of NS Blocks.}
To access the number of normalized self-attention blocks under different settings, we derive three variants as \#7, \#8, and \#9.
We observe that \#8 (\ourmodel~setting) is significantly better than \#7 and \#9, with $R=2$, in all metrics on CVC-300-TV and CVC-612-T.
%
%
This improvement illustrates that too many iterations of NS blocks may cause overfitting on small datasets (\#9).
In contrast, the model fails to alleviate the diffusion issue of high-level features with a single residual block. Empirically, we recommend increasing the number of NS blocks when training on larger datasets.

\section{Conclusion}
We have proposed a self-attention based framework, \ourmodel, 
to accurately segment polyps from colonoscopy videos with super high speed ($\sim$140fps).  
Our basic normalized self-attention blocks can be easily plugged into existing CNN-based architectures. 
We experimentally show that our \ourmodel~achieves the best performance on all existing publicly available datasets under six metrics.
Further, extensive ablation studies demonstrate that the core components in our \ourmodel~are all effective.
We hope that the proposed \ourmodel~can serve as a catalyst for progressing both in VPS as well as other closely related video-based medical segmentation tasks. 
Exploring the performance of \ourmodel~on a larger VPS dataset will be left to our future work.


\bibliographystyle{splncs04}
\bibliography{refs}

\end{document}